\documentclass[11pt]{article}

\usepackage[final]{acl}

\usepackage{times}
\usepackage{latexsym}

\usepackage[T1]{fontenc}
\usepackage[utf8]{inputenc}

\usepackage{microtype}

\usepackage{inconsolata}

\usepackage{graphicx}

\usepackage{xspace}
\usepackage{multirow}
\usepackage{multicol}
\usepackage{booktabs}
\usepackage{amsmath}
\usepackage{amsfonts}
\usepackage{array}
\usepackage{amsthm}
\usepackage{amssymb}
\usepackage{color}
\usepackage{xcolor}
\usepackage{soul}
\usepackage{subfigure}
\usepackage{listings}
\usepackage{utfsym}
\usepackage{colortbl}
\usepackage{longtable}
\usepackage{tabularray}
\usepackage{tabularx}
\usepackage{tcolorbox}
\tcbuselibrary{breakable}
\usepackage{fontawesome5} % For beautiful icons
\usepackage{xcolor} % For color definitions
\tcbuselibrary{skins}
\tcbuselibrary{listingsutf8}

\definecolor{rqblue}{HTML}{E8F1FA}
\definecolor{findinggreen}{HTML}{E8F9E8}

\newtcolorbox{rqbox}[1][]{
    breakable,
    enhanced,
    sharp corners,
    boxrule=0pt,
    colback=rqblue,
    colframe=rqblue,
    frame hidden,
    borderline west={2pt}{0pt}{blue!80!black},
    left=6pt,
    right=6pt,
    top=4pt,
    bottom=4pt,
    before skip=10pt,
    after skip=10pt,
    fontupper=\linespread{1.1}\selectfont,
    #1
}

\newtcolorbox{findingbox}[1][]{
    breakable,
    enhanced,
    sharp corners,
    boxrule=0pt,
    colback=findinggreen,
    colframe=findinggreen,
    frame hidden,
    borderline west={2pt}{0pt}{green!70!black},
    left=6pt,
    right=6pt,
    top=4pt,
    bottom=4pt,
    before skip=10pt,
    after skip=10pt,
    fontupper=\linespread{1.0}\selectfont,
    #1
}

\newcolumntype{L}[1]{>{\raggedright\arraybackslash}p{#1}}

\definecolor{ming}{HTML}{0171C8}

\newtcolorbox{resultbox}{
  breakable,
  sharp corners,
  colframe=black,
  colback=white,
  boxrule=0.4pt,
  left=2pt,
  right=2pt,
  top=2pt,
  bottom=2pt,
  fontupper=\linespread{0.7}\selectfont
}

\newcommand*\samethanks[1][\value{footnote}]{\footnotemark[#1]}
\newcommand{\authorspace}{,\hspace{0.4cm}}

\makeatletter
\renewcommand*{\@fnsymbol}[1]{\ensuremath{\ifcase#1\or \dagger\or \ddagger\or \mathsection\or \mathparagraph\or \|\or **\or \dagger\dagger \or \ddagger\ddagger \else\@ctrerr\fi}}
\makeatother

\newtcolorbox{takeaway}{
    colback=gray!15, 
    colframe=black, 
    boxrule=0.8pt, 
    arc=12pt,
    boxsep=2pt,
    left=6pt, right=6pt, top=6pt, bottom=6pt
}

\title{\textit{Opening the Black Box:} A Survey on the Mechanisms of \\ Multi-Step Reasoning in Large Language Models}

\author{
\bf Liangming Pan$^{1}$\thanks{Corresponding Authors}\authorspace
\bf Jason Liang$^{2}$\authorspace
\bf Jiaran Ye$^{3}$\authorspace\\
\bf Minglai Yang$^{3,4}$\authorspace 
\bf Xinyuan Lu$^{5}$\authorspace
\bf Fengbin Zhu$^{5}$\samethanks\\[0.5em]
$^1$Peking University \quad $^2$Stanford University \quad $^3$Tsinghua University \\
$^4$University of Arizona \quad $^5$National University of Singapore\\[0.5em]
\texttt{liangmingpan@pku.edu.cn}
}

\begin{document}
\maketitle

\begin{abstract}
% Large Language Models (LLMs) have demonstrated remarkable abilities to solve problems requiring multiple reasoning steps, but the mechanisms enabling this multi-step reasoning remain elusive. This survey provides a comprehensive overview of current research into \textit{how} LLMs execute multi-step reasoning, \textit{how} such capabilities are learned, and \textit{how} explicit reasoning strategies like chain-of-thought (CoT) prompting compare to implicit latent reasoning. We distinguish between \textbf{implicit} reasoning (latent multi-step inference hidden in model activations), \textbf{explicit} reasoning (chain-of-thought style step-by-step reasoning in outputs), and \textbf{long-thinking} paradigms that push reasoning to greater lengths. Open research questions in each area are highlighted, with an emphasis on understanding the internal mechanisms, training dynamics, shortcut risks, and explainability of multi-step reasoning in LLMs.

Large Language Models (LLMs) have demonstrated remarkable abilities to solve problems requiring multiple reasoning steps, yet the internal mechanisms enabling such capabilities remain elusive. Unlike existing surveys that primarily focus on engineering methods to enhance performance, this survey provides a comprehensive overview of the \textit{mechanisms} underlying LLM multi-step reasoning. We organize the survey around a conceptual framework comprising seven interconnected research questions from how LLMs execute \textit{implicit multi-hop reasoning} within hidden activations to how \textit{verbalized explicit reasoning} remodels the internal computation. Finally, we highlight five research directions for future mechanistic studies. 
% [Long] Finally, we highlight five under-explored open research questions to outline a roadmap for future mechanistic studies. 
\end{abstract}

% \begin{abstract}
% Large Language Models (LLMs) have demonstrated remarkable abilities to solve problems requiring multiple reasoning steps, yet the internal mechanisms enabling such capabilities remain elusive. Unlike existing surveys that primarily focus on engineering methods to enhance performance, this survey provides a comprehensive overview of the \textit{mechanisms} underlying LLM multi-step reasoning. We organize the survey around a conceptual framework comprising seven interconnected research questions from how LLMs execute \textit{implicit multi-hop reasoning} within hidden activations, to how \textit{verbalized explicit reasoning remodels the internal computation. 
% % latent implicit reasoning to verbalized explicit reasoning. 
% Finally, we highlight five under-explored open research questions to outline a roadmap for future mechanistic studies. 
% \end{abstract}

% how LLMs execute \textit{implicit multi-hop reasoning} within hidden activations, to how \textit{verbalized explicit reasoning remodels the internal computation. 

%%% Introduction
\section{Introduction}
% Background of Multi-Step Reasoning
Large Language Models (LLMs) have demonstrated an impressive ability to carry out \textit{\textbf{multi-step reasoning}}, which involves the process of drawing conclusions through a sequence of intermediate steps, where each step builds on the previous one. Multi-step reasoning has been widely regarded as one of the most fundamental forms of reasoning~\cite{hou_towards_mechanistic_multi_step,how_two_hop_in_context}. It serves as the backbone of advanced tasks such as deep question answering, mathematical problem solving, logical deduction, code generation, and planning~\cite{codex_model,wei2023chainofthought,gpt4_report,qwen25_report,llama3_herd_of_models,deepSeek_coder,deepSeek_R1}.

Multi-step reasoning in LLMs generally takes on two distinct forms. \textbf{\textit{Implicit reasoning}} involves performing multi-hop inference entirely within the model's hidden activations, delivering a correct final answer without verbalizing intermediate steps. In contrast, \textbf{\textit{explicit reasoning}}, exemplified by \textit{Chain-of-Thought} (CoT)~\cite{wei2023chainofthought}, instructs the model to externalize the reasoning process into a sequence of natural language tokens. 
% ,zeroshot-cot
% After extensive pretraining, models often exhibit an emergent ability of \textbf{\textit{implicit multi-step reasoning}}, \textit{i.e.}, producing the correct answer without explicitly verbalizing intermediate steps. 
Remarkably, modern LLMs have exhibited strong performance in both paradigms~\cite{cot_survey_24,long_cot_survey,implicit_reasoning_survey}. Building on this empirical success, the \textbf{\textit{internal mechanisms}} that enable such capabilities become scientifically intriguing.
For implicit reasoning, a key puzzle is \textit{how} multi-step reasoning capabilities emerge from simple next-token prediction training, and \textit{how} LLMs internally carry out multi-step computations. 
% For explicit CoT reasoning, critical questions persist about the causal role of the generated rationale: why CoT can elicit superior reasoning capabilities and does the generated rationale faithfully reflect the model's actual decision-making process. 
For explicit CoT reasoning, critical questions persist about \textit{why} CoT can elicit superior reasoning capabilities and \textit{whether} the generated rationale faithfully reflects the model's actual decision-making process. Understanding these mechanisms is not only a matter of scientific curiosity but also a prerequisite for building more reliable, controllable, and human-aligned reasoning systems. 
% but also a prerequisite for building reasoning systems that are more reliable, controllable, and aligned with human intentions. 

\begin{figure*}[!t]
	\centering
	\includegraphics[width=16cm]{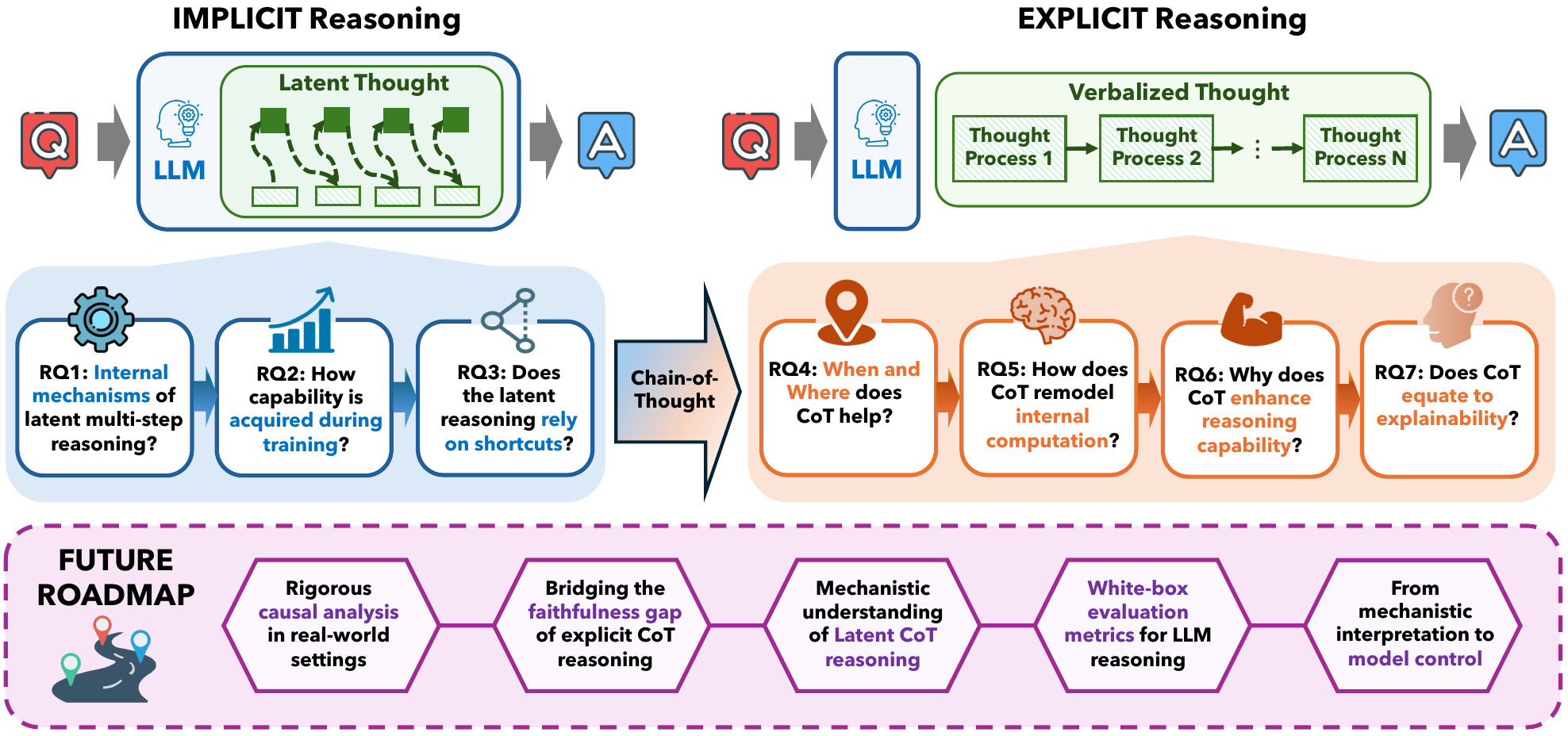}
    \caption{The cognitive framework and organizational structure of this survey. We explore the mechanisms of multi-step reasoning through two distinct paradigms: \textit{Implicit Reasoning} and \textit{Explicit Reasoning}, through seven interconnected \textit{Research Questions}. The bottom panel highlights five strategic directions for future research. 
    }
    \vspace{-0.2cm}
    \label{fig:general_framework}
\end{figure*}

% phenomena
% Although we still lack a definitive theory for all these scientific questions, 
Although we still lack a unified mechanistic theory, a growing body of literature seeks to \textit{open the black box of LLM multi-step reasoning} and has made significant progress. In this paper, we aim to provide a comprehensive overview of these works. Unlike existing surveys~\cite{reasoning_survey_huang,cot_24_survey,long_cot_survey} that primarily focus on \textit{enhancing} reasoning  (\textit{e.g.}, through tool use, retrieval augmentation, or self-correction), our survey explicitly focuses on \textit{\textbf{understanding mechanisms}}, a perspective that has been largely overlooked in previous reviews. As illustrated in Figure~\ref{fig:general_framework}, we identify seven pivotal, interconnected, and progressive \textit{research questions (RQs)} to form the cognitive framework of our survey. These questions form a cohesive narrative, covering analytical methods and key findings from the hidden internal dynamics of latent reasoning to the visible mechanisms of explicit CoT reasoning. 
% guiding readers from the hidden internal dynamics of latent reasoning to the visible mechanisms of explicit CoT reasoning. 
We end by pointing out five open research questions that remain under-explored but are essential for the future roadmap of mechanistic understanding. 
% in LLM reasoning. 

% We end by looking beyond existing works to point out five open research questions that remain under-explored but are essential for the future roadmap of mechanistic understanding in LLM reasoning. 

% In this paper, we aim to provide a comprehensive overview of research that seeks to \textit{uncover the mechanisms underlying LLM multi-step reasoning}. Although several surveys on LLM reasoning exist, they primarily focus on \textit{enhancing} reasoning --- through tool use, retrieval augmentation, or self-correction~\cite{reasoning_survey_huang,cot_24_survey,long_cot_survey} --- rather than on \textit{understanding} the mechanisms behind it. This survey aims to fill that gap. We systematically review progress on the mechanisms of multi-step reasoning, organizing the literature by the research problems posed, the analytical methods employed, and the key findings obtained. 

% To provide a structured and comprehensive view of how reasoning emerges and functions, we identify seven pivotal, interconnected, and progressive research questions (RQs), illustrated in Figure 1 as the cognitive framework of our survey. 

%%% Main Figure

%%% Implicit Multi-Step Reasoning
\section{Implicit Multi-Step Reasoning}
% \subsection{Internal Mechanisms of Latent Multi-Hop Reasoning}

Multi-hop implicit reasoning is the process of answering a question by combining multiple pieces of information across several steps. Unlike explicit reasoning, the intermediate links are not directly stated and must be inferred using background knowledge or context. Mechanistic study of multi-hop implicit reasoning is important because it reveals whether models truly perform step-by-step reasoning or rely on shallow shortcuts. Such understanding improves interpretability and trust in LLMs, and it guides the development of models that generalize more reliably. 

\subsection{What are the internal mechanisms of latent multi-step reasoning?}
\label{subsec:implicit:internal_mechanism}
% interpretability
Recent mechanistic studies have begun to unveil how LLMs carry out latent multi-hop computation entirely in their hidden states~\cite{yang2024large,biran2024hopping,DBLP:conf/acl/BrinkmannSLSB24}. These studies employing causal probing, mechanistic tracing, and representational analysis have collectively revealed a \textit{staged} internal process in which intermediate results are computed and transformed layer by layer, ultimately contributing to the final output. In essence, transformers appear to implement an internal chain-of-thought spread across their depth.

% chain-of-thought

% Recent studies employing causal probing, mechanistic tracing, and representational analysis have advanced a nuanced understanding of this phenomenon, wherein intermediate results are computed internally but not explicitly produced. 

% \paragraph{Layer specification and information flow.}
\paragraph{Functional specialization of layers.}
% A major body of work aims to identify how different transformer layers take on distinct roles in a multi-hop inference. 
A major body of work explores \textit{layer specialization}, aiming to identify the distinct computational roles each layer plays during multi-hop inference. 
Using \textit{Patchscopes}~\cite{DBLP:conf/icml/GhandehariounCP24} together with a novel intervention technique termed \textit{back-patching}, \citet{biran2024hopping} uncovered a sequential computation pathway in which early layers identify the bridge entity, which is then propagated forward and exploited by later layers to complete the inference. Complementarily, \citet{li2024understanding} applied logit lens analysis~\cite{nostalgebraist2020logitlens} and found that implicit reasoning representations emerge in intermediate layers and have a causal influence on generating the final answer. 
% Complementarily, \citet{li2024understanding}, through Logit Lens analysis~\cite{nostalgebraist2020logitlens} and intervention experiments, showed that implicit reasoning representations emerge in intermediate layers and exert a causal influence on the generation of final answers. 
Extending this perspective, \citet{yu2025back} traced logits through the network via a neuron-level logit flow method and observed that even a single-hop query is solved in multiple distinct stages---entity subject enrichment, entity attribute extraction, relation subject enrichment, and relation attribute extraction---each of which is localized to different layers. More recently, \citet{DBLP:journals/corr/abs-2505-14530} showed that this layer-wise reasoning also applies at the task level: for composite instructions, models execute different subtasks at different depths, forming a staged computation across layers. All the above studies provided evidence of functional specialization of transformer layers in multi-hop reasoning. 

% an implicit decomposition of the task across the transformer's depth in multi-hop reasoning. 
%This kind of functional specialization suggests an implicit decomposition of the task across the transformer’s depth. 
%the functional specialization of transformer layers in 
%This kind of functional specialization suggests an implicit decomposition of the task across the transformer's depth in multi-hop reasoning. 

% \paragraph{Trees as a structural lens for reasoning.} 
\paragraph{Uncovering fine-grained reasoning structures.}
Beyond layer specification, another line of work aims to recover more fine-grained implicit reasoning structures from model internals. \textit{MechanisticProbe}~\cite{hou2023mechanisticprobe} introduced an attention-probing technique to extract latent reasoning trees from transformer activations. They showed that on synthetic and natural tasks with GPT-2 and LLaMA, models often perform procedural reasoning layer by layer, with lower layers selecting statements and higher layers executing reasoning steps. Complementing these findings, \citet{brinkmann2024mechanisticanalysis} analyzed a small transformer trained on a symbolic tree path-finding task, finding that it implements a backward chaining algorithm: deduction heads climb trees one level per layer, register tokens act as working memory for parallel subpaths, and a one-step lookahead heuristic compensates when chaining is insufficient. Together, these studies demonstrate that transformers can adopt structured, algorithm-like reasoning strategies beyond memorization, albeit within the limits of the model's depth (to be discussed below). 
% (more on this below). 

% albeit within architectural depth limits. 
% These studies demonstrate that transformers can learn algorithm-like, structured reasoning strategies beyond mere memorization, albeit within the limits of the model's architecture (more on this below). 

\paragraph{Layer depth as the primary bottleneck for implicit reasoning.}
% \paragraph{Layer depth as the driving factor for implicit reasoning.}
% Layer depth acts as a computational bottleneck for reasoning. 
Theoretical and empirical studies indicate that the number of reasoning steps a model can perform implicitly is strictly limited by its depth. \citet{DBLP:conf/iclr/MerrillS24} theoretically demonstrated that a standard Transformer with constant depth cannot solve inherently serial problems that require computation scaling with input size, \textit{e.g.}, parity or graph connectivity. In practice, \citet{yu2025llms} and \citet{how_two_hop_in_context} found that specific multi-hop reasoning tasks require a minimum threshold of layers to resolve; if a model is too shallow, the ``latent chain'' is cut short, and the reasoning fails. \citet{saunshi_reasoning} formally established that an $L$-layer Transformer can simulate an $m$-step explicit reasoning process, provided $L$ is sufficiently large to accommodate the iterative forward passes required. All these works revealed a close correlation between layer depth and the implicit reasoning capabilities of the model. 

\paragraph{Why implicit reasoning sometimes fails.} 
Identifying how and why implicit reasoning sometimes fails has also been illuminating. 
% One of the most important aspects of the mechanistic analysis of LLMs is to understand why latent reasoning sometimes fails. 
\citet{biran2024hopping} discovered that many failures stem from delayed resolution of the first hop, and showed that rerunning computations via back-patching can correct these errors.
% , thereby exposing architectural limitations in transformer-based models. 
\citet{li2024understanding} found that failures frequently arise from the improper generation or utilization of implicit reasoning results. To address this, they proposed \textsc{CREME}, a lightweight model-editing technique that patches specific multi-head self-attention modules, leading to improved compositional reasoning generalization with minimal disruption to unrelated predictions. 
% To address this, they proposed CREME, a lightweight model-editing technique that patches specific attention heads, leading to improved compositional reasoning with minimal disruption to unrelated predictions. 
In the context of two-hop queries (``$e_{1}$'s $r_{1}$'s $r_{2}$ is $e_{3}$''), \citet{yu2025back} showed that errors often occur when high-layer features at the $r_{1}$ position overemphasize the intermediate entity $e_{2}$, outweighing the logits for the correct final entity $e_{3}$. This finding revealed that LLMs internally build and combine entity--relation representations in a staged manner, but positional interference can derail multi-hop reasoning. 
% In two-hop QA, \citet{yu2025back} discovered a common failure mode: higher layers often over-emphasize the intermediate entity at the expense of the correct final answer. 
To fix this, they introduced a back-attention mechanism allowing lower layers to reuse higher-layer information from other positions, which substantially improved multi-hop accuracy. 
% To mitigate this issue, they propose \textit{back attention}, which allows lower layers to access higher-layer states from other positions, thereby restoring crucial features and substantially improving multi-hop reasoning accuracy in both trained-from-scratch transformers and pretrained LLMs. 
However, even with such interventions, certain transformers still struggle to reliably chain more than one reasoning step. For example, \citet{yang2024large} found that \textit{LLaMA-2} models, while reliably recalling a needed bridge entity, often fail to apply it to the second hop, highlighting limits in architecture that impede consistent multi-step chaining. 
% \citet{yang2024large} use mechanistic probes—\textit{entity recall} and \textit{consistency} scores—to show that while LLaMA-2 reliably recalls bridge entities for the first hop, it only inconsistently applies them for the second. This breakdown highlights architectural limits that hinder reliable chaining of reasoning steps. 

% \paragraph{Evidence for distributional reasoning.} 
% This indicates that transformers might internally consider several potential reasoning paths in parallel, before converging on the answer. 
% Extending prior insights into multi-hop reasoning in LLMs, \citet{shalev2024distributional} introduce the concept of \textit{distributional reasoning}, wherein intermediate layers encode a probability distribution over multiple plausible intermediate answers. This representation gradually sharpens through a \textit{phase transition}, which suppresses less likely candidates while amplifying the correct one, thereby transforming a diffuse distribution into a discrete final prediction. 
% % —often mediated by feed-forward networks (FFNs)—
% Beyond discrete intermediate steps, researchers have also found evidence for distributional internal reasoning. 

\begin{takeaway}
    \textbf{Takeaway:}
    Implicit multi-hop reasoning in LLMs is not a monolithic capability but rather an orchestrated, layered process, with distinct modules and pathways specializing in different phases of the reasoning chain. 
    % Probing and intervention studies showed that intermediate results, \textit{e.g.}, bridge entities, are computed and passed along inside the network. 
    For example, probing and intervention studies showed that intermediate results, \textit{e.g.}, bridge entities, are computed and passed along inside the network. 
    Nevertheless, such implicit reasoning is constrained by the inherent architecture of transformers, for example, their fixed depth. 
    % Mystery remains in the failure mode 
    % To address these limitations, a growing body of work has proposed techniques aimed at enhancing performance. 
\end{takeaway}

\subsection{How latent multi-step reasoning capability is acquired during training?}
\label{subsec:implicit:learning}
Models do not possess latent reasoning capabilities at initialization. If multi-hop reasoning is implemented via specialized internal circuits discussed in Section~\ref{subsec:implicit:internal_mechanism}, a critical question arises: \textit{how do these circuits emerge in the first place?} Research into training dynamics reveals that implicit reasoning is an \textit{acquired} behavior that emerges during the training process through distinct phase transitions. 
% they are \textit{acquired} behaviors that emerge during the training process. 

\paragraph{Grokking marks the shift from memorization to reasoning.} 
Recent studies~\cite{wang2024grokked,ye2025does,zhang2025complexity,abramov_grokking} suggested that LLMs do not learn multi-step reasoning gradually; instead, they often undergo \textit{phase transitions} during training where reasoning capabilities appear \textit{suddenly} rather than continuously. In other words, a model might spend many updates seemingly memorizing or floundering, then ``grok'' the underlying reasoning algorithm after a certain point. This phenomenon, known as \textit{``grokking''}, was initially observed in deep networks trained on other tasks such as modular arithmetic, where generalization performance spikes long after training accuracy has saturated~\cite{power2022grokking,olsson2022context,wei2022emergent}. 

In the context of multi-hop implicit reasoning, this phenomenon of transformers transitioning from early-stage \textit{memorization} to later-stage \textit{generalization} was first observed by \citet{wang2024grokked} through training transformers from scratch on symbolic reasoning tasks. They found that the multi-hop reasoning capability emerges only through \textit{grokking}, where an early \textit{memorizing circuit} is gradually replaced by a more efficient \textit{generalizing circuit} due to optimization bias and weight decay. \citet{ye2025does} corroborated this phase transition, proposing a \textit{three-stage trajectory}: (i) rapid memorization, (ii) delayed in-distribution generalization, and (iii) slower cross-distribution generalization, with persistent OOD bottlenecks at the second hop. Mechanistically, they employed \textit{cross-query semantic patching} to localize the ``bridge'' entity and a \textit{cosine-based representational lens} to reveal that generalization coincides with \textit{mid-layer clustering} of intermediate entity representations.

% Wang et al. (2024b) trained transformers from scratch on synthetic reasoning tasks and observed that multi-hop reasoning only emerged via grokking – that is, after the model had already memorized the training set and reached 100\% training accuracy. At that critical juncture, the model began to replace its earlier memorization-based circuits with new, more generalizable ``reasoning” circuits''. 

% Previous research \cite{olsson2022context, wei2022emergent} has observed an interesting phenomenon: modern deep learning models often undergo sudden shifts in their internal computational structure, accompanied by changes in input–output behavior. In particular, \citet{hooglandloss} shows that the training process of transformers can be divided into distinct stages, each marked by key transitions in both their internal organization and external performance. Similar to the earlier observation, the studies---\citet{wang2024grokked}, \citet{ye2025does}, and \citet{zhang2025complexity}---discovered a \textit{phase transition} in which training dynamics shift from early-stage \textit{memorization} to later-stage \textit{generalization} for multi-step reasoning tasks. This transition arises only after the model attains perfect in-distribution training accuracy and involves the gradual replacement or augmentation of memorization circuits with reasoning-oriented mechanisms. 

% \paragraph{Factors that steer this phase transition.} 
% \paragraph{Factors influencing the emergence of reasoning circuits.}
\paragraph{Factors influencing the emergence of reasoning.} 
The transition from memorization to generalization is not random; studies revealed that it is governed by specific properties. One of the primary determinants is the \textit{training data distribution}. \citet{wang2024grokked} demonstrated that the speed of grokking correlates strongly with the \textit{ratio of inferred to atomic facts $\phi$} in training. A higher ratio of compositional examples forces the model to abandon inefficient memorization in favor of the generalizing circuit. Expanding this to real-world scenarios, \citet{abramov_grokking} found that natural corpora often lack sufficient connectivity (low $\phi$) to trigger this transition, but data augmentation with synthetic inferred facts can artificially raise $\phi$ above the critical threshold required for circuit formation. 
% Beyond data distribution, \textit{complexity control} also steers the phase transition. % formation of reasoning circuits. 
% \citet{zhang2025complexity} showed that smaller initialization scales and stronger weight decay bias the optimization process toward \textit{low-complexity, rule-like solutions} rather than high-complexity, memory-based mappings, thereby accelerating the emergence of reasoning capabilities. 
Beyond data distribution, the \textit{scale of the training data} also matters. \citet{yaoimplicitreasoning} revealed a scaling law: the data budget required to learn implicit $k$-hop reasoning grows exponentially with $k$, though curriculum learning can significantly mitigate this cost. From an optimization perspective, \citet{zhang2025complexity} identified \textit{complexity control} parameters as crucial factors. They found that smaller initialization scales and stronger weight decay bias the optimization process toward low-complexity, rule-like solutions rather than high-complexity, memory-based mappings, thereby accelerating the emergence of reasoning capabilities. Finally, \citet{li_where_to_find_grokking} observed that in large-scale pretraining, grokking is \textit{asynchronous and local}; different domains and data groups undergo this memorization-to-generalization transition at different times depending on their inherent difficulty and distribution heterogeneity.

% Beyond data distribution, the \textit{complexity of the reasoning task} imposes strict scaling laws: \citet{yaoimplicitreasoning} showed that the data budget required to learn implicit $k$-hop reasoning grows exponentially with $k$, though curriculum learning can significantly mitigate this cost. From an optimization perspective, \citet{zhang2025complexity} identified \textit{complexity control} parameters, such as initialization scale and weight decay, as crucial levers. They found that smaller initialization scales and stronger weight decay bias the model toward low-complexity, rule-based solutions, thereby accelerating the reasoning phase. Finally, \citet{li_where_to_find_grokking} provided a realistic nuance, observing that in large-scale pretraining, grokking is \textit{asynchronous and local}; different domains and data groups undergo this memorization-to-generalization transition at different times depending on their inherent difficulty and distribution heterogeneity. 

\begin{takeaway}
    \textbf{Takeaway:}
    Implicit multi-hop reasoning capability is an \textit{acquired} capability that emerges via \textit{grokking}---a phase transition from surface-level memorization to structured reasoning. This transition is not automatic; it is governed by critical factors, including the training data distribution, the data scale, and complexity control via optimization biases. 
\end{takeaway}

\subsection{To what extent does multi-step reasoning rely on shortcuts?}
\label{subsec:implicit:shortcuts}
While the training dynamics discussed in \S~\ref{subsec:implicit:internal_mechanism} suggest that structured reasoning circuits can emerge, growing mechanistic evidence has also uncovered a more complex and often discouraging reality regarding model internals. 
% a growing body of mechanistic evidence reveals a more complex reality. 
% ~\cite{yang2024large_shortcut, lin2025implicitreasoning, ju2024investigating, guo2025llms} 
Models frequently bypass genuine multi-step reasoning, relying instead on \textit{``shortcuts''}---statistical correlations or surface-level heuristics that mimic reasoning without performing the underlying computation~\cite{kang2023impact,elazar2023s,DBLP:conf/acl/YangKG0G25}. 

\paragraph{Factual shortcuts bypass intermediate reasoning.} 
A primary form of shortcutting involves exploiting direct associations between the subject and the final answer, effectively skipping the intermediate steps. \citet{ju2024investigating} investigated this in the context of knowledge editing, finding that failures often stem from ``shortcut neurons'' that encode a direct link between the first and last entities, ignoring the multi-hop structure. Mechanistically, \citet{yang2024large_shortcut} used \textit{Patchscopes}~\cite{DBLP:conf/icml/GhandehariounCP24} to distinguish valid reasoning from shortcuts. They observed that genuine implicit reasoning coincides with the model constructing a hidden representation of the intermediate bridge entity. In contrast, shortcut-prone queries bypass this internal construction entirely. When these direct shortcuts are removed, model performance drops by nearly a factor of three, revealing that much of the perceived reasoning capability is illusory. 

\paragraph{Shortcuts based on surface-level pattern matching.} 
Beyond factual associations, models also latch onto structural regularities in the training data. \citet{lin2025implicitreasoning} analyzed implicit arithmetic reasoning and found that models often adopt a ``bag-of-words'' heuristic, treating operations as commutative even when they are not. While this shortcut works for fixed-template examples, performance collapses when premise order is randomized, proving the model had not learned the robust sequential logic. Similarly, \citet{how_two_hop_in_context} found that in the presence of context distractors, pretrained models default to a heuristic of guessing based on surface plausibility. However, they also noted a positive trajectory: fine-tuning can force a phase transition where the model shifts from this shallow guessing behavior to a sequential query mechanism that explicitly retrieves intermediate entities.

\begin{takeaway}
    \textbf{Takeaway:}
    LLMs frequently bypass the ``latent reasoning chain'' via \textit{factual shortcuts} (direct input-output associations) or \textit{structural heuristics} (exploiting surface patterns like commutativity). This underscores the need for shortcut-free evaluation protocols and training setups that force models to construct and reuse intermediate representations. 
\end{takeaway}

% Taken together, these findings show that even when models succeed on multi-hop or multi-step reasoning benchmarks, much of that success can stem from \textit{heuristic or distributional shortcuts}---co-occurrence patterns in factual domains, or exploitable algebraic regularities in mathematical domains---rather than robust, generalizable reasoning mechanisms. This underscores the need for shortcut-free evaluation protocols (as in SOCRATES) and training setups that force models to construct and reuse intermediate representations even under adversarial input conditions.

% Much of the internal working of implicit multi-step reasoning relies on heuristic or distributional shortcuts

% LLMs frequently bypass the ``latent reasoning chain'' via factual shortcuts (direct input-output associations) or structural heuristics (exploiting surface patterns like commutativity). True implicit reasoning involves constructing intermediate representations (e.g., bridge entities), but this is often circumvented by statistical correlations. Future evaluation must strictly control for shortcuts to measure true compositional ability. 

%%% Explicit Multi-Step Reasoning
\section{Explicit Multi-Step Reasoning}
% Prompting an LLM to produce a step-by-step \textit{chain-of-thought} (CoT) has been shown to unlock significantly better performance on tasks that require reasoning. Chain-of-thought prompting involves asking the model to first generate a sequence of intermediate reasoning steps (often in natural language) and then give the final answer. 

% Implicit multi-step reasoning operates entirely within the fixed computational budget of the model's hidden states; therefore, it is bounded by the depth bottleneck or falls prey to shortcuts. \textit{Explicit multi-step reasoning} fundamentally alters this paradigm, where the reasoning process is externalized via Chain-of-Thought (CoT). 

Implicit reasoning operates entirely within the fixed computational budget of the model's hidden states; therefore, it is bounded by the depth bottleneck and frequently falls prey to shortcuts. \textit{Explicit multi-step reasoning} fundamentally alters this paradigm. By prompting an LLM to produce a step-by-step \textit{Chain-of-Thought} (CoT), the reasoning process is externalized into a sequence of natural language tokens, effectively extending the computational capacity beyond the model's layers. CoT has been shown to unlock significantly better performance on tasks that require reasoning. In this section, we dissect the mechanisms of this paradigm through four progressive research questions (\S~\ref{subsec:explicit:whenandwhere}-\S~\ref{subsec:explicit:cot_explanation}).  
% : determining when and where CoT is effective (RQ4), analyzing how it remodels internal computation (RQ5), explaining why it enhances reasoning capabilities (RQ6), and evaluating whether these explicit rationales faithfully reflect the model's actual decision-making (RQ7).

% Empirical studies have found that CoT is especially beneficial for tasks in arithmetic, logical reasoning, commonsense reasoning, and other problems that humans would naturally solve via intermediate steps~\cite{wei2023chainofthought}. 

\subsection{Where and When Does CoT Help?}
\label{subsec:explicit:whenandwhere}

\paragraph{On which tasks does CoT help?} 
To uncover this, \citet{to_cot_or_not} conducted a large-scale meta-analysis across 20 benchmarks and found that prompting with CoT yields large gains primarily on \textit{math and symbolic logic tasks}, with far smaller or even negative gains on other domains. \citet{Challenging_BIG_Bench} similarly showed that many \textit{BIG-Bench Hard} tasks~\cite{Big_Bench}, which had stumped standard few-shot prompts, become solvable with CoT. These were precisely tasks requiring multi-step reasoning, \textit{e.g.}, symbolic manipulation, compositional logic. However, for knowledge-heavy tasks like MMLU~\cite{MMMU} or commonsense reasoning, CoT often provides negligible improvement~\cite{to_cot_or_not}. 
% While CoT generally aids reasoning, it is not a universal remedy. 
In certain cases, CoT can even degrade accuracy. 
% In certain cases where the CoT is unnecessary, generating a chain-of-thought can degrade accuracy. 
For example, \citet{Mind_your_step} examined cognitive‐psychology tasks where additional deliberation harms human performance, \textit{e.g.}, certain trick riddles or intuitive judgment problems. They found that CoT substantially degraded accuracy on such tasks, and it tends to distract the model into over-complicating a problem that might have been solved via intuition. A complementary study on Blocksworld planning~\cite{chain_of_thoughtlessness} found that CoT helps only when the prompt examples closely match the test distribution, and the gains quickly deteriorate if the test problem's complexity exceeds that seen in the exemplars. 

% \researchfinding{CoT’s benefits are brittle: it can \emph{backfire} on intuition-dominant tasks and degrade sharply outside the prompt's demonstrated scope.} 

%%%%%%%%%%
% FACTORS INFLUENCING COTs!!!
% Empirical studies have shown that CoT performance can be dramatically influenced by many features of the CoT prompt (Madaan and Yazdanbakhsh, 2022; Wu et al., 2023; Jin et al., 2024); for example, the relevance and ordering of reasoning fragments is more important
% than their accuracy (Wang et al., 2023a,b; Ye and Durrett, 2022), and minor input perturbations can substantially bias models’ answers (Turpin et al., 2024), suggesting that the models lack general reasoning abilities (Stechly et al., 2024). 

% Deciphering the Factors Influencing the Efficacy of Chain-of-Thought:
% Probability, Memorization, and Noisy Reasoning
%%%%%%%%%%

\paragraph{What factors influence the efficacy of CoT?}
% \paragraph{Which parts in the prompt are crucial for CoT?} 
% \paragraph{Which parts of the CoT prompt matter most?} 
% Beyond task-level evaluations of CoT's effectiveness, a complementary line of work examines \textit{which components of the CoT prompt most strongly govern its effectiveness}. 
Beyond task-level evaluations, empirical studies have shown that CoT performance can be dramatically influenced by many features of the CoT prompt. First, studies~\cite{ye2022unreliability,what_makes_cot_effective,empiricalstudy2022cot} reveal that the \textit{relevance and ordering} of exemplars matter more than their semantics; models can still derive correct answers from invalid rationales if the prompt maintains a coherent structure. Second, the length of reasoning is another critical factor, with \citet{jin2024impact} identifying that the number of reasoning steps significantly modulates model performance. Finally, CoT is surprisingly sensitive to phrasing; minor input perturbations can substantially bias models' answers~\cite{llm_dont_always_say,unzipping_black_box}.

\paragraph{Why do these factors influence CoT efficacy?}
% To take one step further, theoretical frameworks were proposed to provide a better understanding of the above phenomena. 
% To explain why these specific prompt features drive performance, recent research provides theoretical and mechanistic groundings. 
% To explain why these specific factors drive CoT performance, several theoretical and mechanistic frameworks have been proposed. 
% To explain why these specific factors drive CoT performance, several works provided theoretical and mechanistic groundings. 
To explain the mechanisms underlying these factors, recent research provided theoretical and mechanistic groundings. 
\citet{tutunov2024why} proposed that CoT efficacy stems from the model's ability to approximate the true conditional distribution of reasoning, where structured exemplars help the model infer the task's latent logic and reduce generation ambiguity. \citet{prabhakar2024deciphering} refined this view through a controlled case study, characterizing CoT as a probabilistic process heavily modulated by output \textit{probability}, task \textit{memorization} in training data, and step-wise \textit{complexity}. Mechanistically, \citet{wu2023analyzing} revealed how specific components of the CoT prompt drive model generation via gradient-based feature attribution. 

\begin{takeaway}
    \textbf{Takeaway:}
    CoT prompting yields significant gains primarily in tasks involving \textit{mathematical, logical, or symbolic reasoning}. Its efficacy depends more on the structural coherence and relevance of exemplars, the length of reasoning, and the prompt phrasing. Several theoretical and mechanistic frameworks were proposed to understand such driving factors. 
\end{takeaway}

\subsection{How Does Chain-of-Thought Remodel Internal Computation?}
\label{subsec:explicit:shortcuts}

%What is the latent algorithmic process of CoT inside the LLM? 
% Uncovering the latent algorithmic process of CoT inside the LLM. 
% How LLMs represent and update state information during CoT reasoning?
% Neuron-level analysis for fine-grained mechanisms of CoT

% Chain-of-thought prompting does more than change an LLM's output format; growing evidence shows that it fundamentally changes the model's \textit{internal computation} [CITES]. 

Chain-of-thought prompting does more than just alter an LLM's output format. 
% When a model switches to CoT mode, 
Growing evidence shows that it fundamentally changes the model's internal computation into a ``reasoning mode'', where the model retrieves and updates information in a stepwise fashion, leveraging the intermediate computational steps as external memory. 

% When a model switches to CoT mode, growing evidence shows that it fundamentally changes the model's \textit{internal computation} into a ``reasoning mode'' that retrieves and updates information in a stepwise fashion, leveraging the intermediate computational steps as external memories. 

\paragraph{The emergence of iteration heads.} First, \citet{Iteration_Head} identified the ``iteration head'' --- an attention head that emerges during CoT. These heads explicitly focus on the model's previously generated tokens to carry forward interim results. For example, in a loop counter task, an iteration head attends to the token ``Step 4'' to generate ``Step 5''. This effectively allows the model to create a virtual \textit{recurrent neural network} (RNN) where the hidden state is externalized as text. In another study of a \textit{Llama-2} model~\cite{LLAMA2} solving multi-step ontology queries, \citet{mechanistic_cot} also identified early-layer attention heads that ``move information along ontological relationships'' in the contexts that are relevant to the current sub-problem. The emergence of iteration heads provides supporting evidence that CoT enables the model to internally utilize generated text as an external memory for sequential reasoning. 

% CoT alters the model's internal computation, enabling it to utilize generated text as an external memory for sequential reasoning.
% that CoT enables the model to internally retrieve and leverage intermediate results for sequential reasoning. 
% Such attention-based mechanisms validate the idea that the model can internally retrieve and leverage intermediate results for sequential reasoning. 

% Rai and Yao (2024) probed the Feed-Forward Layers (FFNs) during CoT and discovered Reasoning Neurons. These neurons exhibit selective activation: they are largely silent during standard text generation but fire specifically to maintain partial results during reasoning steps (e.g., carrying a digit in addition, or holding a variable value). This confirms that CoT recruits specific sub-networks dedicated to variable storage and manipulation, physically distinct from the networks used for language modeling.

\paragraph{Evidence of state maintenance and update.} 
Besides the access to external memory, studies show that LLMs with CoT can also maintain and update dynamic internal states to track the reasoning process. 
% With an external memory, studies further show that LLMs with CoT can effectively encode, maintain, and update intermediate states to track the reasoning process. 
% Another essential aspect of multi-step reasoning is maintaining and updating intermediate states, a form of working memory that tracks the reasoning progress. Growing evidence indicates that transformers with CoT effectively encode such states internally. 
\citet{FSA_COT} found that when using CoT for state-tracking tasks, LLMs embed an implicit finite state automaton in their hidden layers. Specific feed-forward neurons in later layers were found to correspond directly to discrete problem states, forming a circuit that reliably updates with each new reasoning step. This internal state representation is highly robust and works correctly even with noisy or incomplete CoT steps, suggesting the model learns a resilient state-updating algorithm. By probing individual neurons of LLMs, \citet{DBLP:conf/acl/RaiY24} offered more granular evidence of state maintenance. They identified specific ``reasoning neurons'' in Llama-2's feed-forward layers that activate to hold partial results, such as carried values during arithmetic. Their activation helps explain why including particular steps (\textit{e.g.}, an explicit breakdown of a sum) in the CoT prompt is effective: they reliably trigger the neurons responsible for maintaining the intermediate state. 

% \researchfinding{CoT induces models to internally preserve an evolving solution state, and iteratively update that state as reasoning unfolds.}

% They identified specific ``reasoning neurons'' in Llama-2's feed-forward layers that become active during arithmetic CoT tasks. These neurons only fire when certain sub-calculations or carried values are present, and their activation was found to explain why including particular steps in the CoT prompt. 

% This concept of an internal ``scratchpad'' is reinforced by other findings. As previously noted, experiments with hidden or non-semantic CoT (e.g., chains of dots) show that the model's latent representations still carry the information of the missing steps. This implies that the structure of the chain provides computational "space" for the model to store and refine an implicit state, even without human-readable content. 

% CoT induces models to preserve and iteratively update an evolving solution state, a function implemented through mechanisms ranging from high-level state-tracking circuits down to specialized individual neurons. 

% CoT-prompted models internally preserve an evolving solution state, whether distributed across many neurons or concentrated in special ones, and iteratively update that state as reasoning unfolds.

% \paragraph{Extra hidden computation matters more than token semantics.}
% \paragraph{Computational depth outweighs token semantics.}
\paragraph{Computational depth matters more than token semantics.}
Notably, the internal process of sequential reasoning appears to persist even when the CoT rationale lacks semantic meaning. 
% Notably, even when CoT rationales are hidden or nonsensical, this internal process of state updating and sequential reasoning still persists. 
For example, \citet{think_doy_by_dot} replaced the meaningful CoT text with filler tokens (\textit{e.g.}, ``...''). Surprisingly, models could still solve complex reasoning tasks simply by generating these dots. Similarly, \citet{DBLP:conf/iclr/GoyalJRMKN24} found that introducing a learnable ``pause'' token significantly boosts performance on tasks from QA to math. These findings suggest that the semantic content of reasoning steps may be secondary to the computational time they buy. The sheer act of generating extra tokens (regardless of their meaning) provides necessary computational depth; each token grants the model an additional forward pass through all its layers. This extra ``think time'' enables the model to implement complex reasoning algorithms that cannot be executed in a single pass. \citet{understanding_hidden_computations_cot} reinforced this interpretation through a mechanistic study. They found that even when CoT steps are replaced by placeholders, the model's deeper layers still encode the missing steps, which can be recovered to their correct semantic content via a logit lens probe. 

\paragraph{Parallelism and reasoning shortcuts.}
Finally, although growing evidence reveals the sequential nature of CoT's internal computation, other studies have found that \textit{LLMs often run multiple reasoning pathways in parallel during CoT}, meaning that the model's internal reasoning process is not strictly sequential. 
% in the way the external chain-of-thought is. 
For example, \citet{mechanistic_cot} identified a ``functional rift'' where the model simultaneously tries to solve the problem directly from the question (``reasoning shortcuts'') while also following the step-by-step procedure, and these parallel approaches then converge in later layers. \citet{Arithmetic_Without_Algorithms} found that models perform arithmetic via a ``bag of heuristics'' (many simple feature detectors) rather than a single step-by-step algorithm. \citet{cot_in_the_wild_not_faithful} observed that the models can still arrive at the correct answer, even if they might make a mistake in an early step internally. 
% These studies provide counter-evidence to a faithful step-by-step execution inside the model. 
The above evidence on parallelism and shortcuts reveals that CoT's internal workings are more complicated. It is a combination of sequential step-by-step reasoning, parallel associative shortcuts, and occasional after-the-fact rationalizations. 

% Therefore, while some studies have identified circuit-like components (\textit{e.g.}, heads that track state or retrieve facts) that support human-like, step-by-step reasoning, evidence on parallelism and shortcuts reveals that CoT's internal workings are more complicated. It is a combination of sequential algorithmic steps with parallel shortcuts and after-the-fact rationalizations. 

% In summary, CoT prompting causes LLMs to activate information-retrieval circuits at each step – often embodied by attention heads that pull in the necessary facts or previous outputs – thereby mimicking how a human reasoner recalls facts or prior results throughout a multi-step solution.

% CoT activates stepwise retrieval circuits, \textit{i.e.}, early heads fetch relevant facts and carry forward interim results. This mimics how a human reasoner recalls facts or prior results throughout a multi-step solution. This behavior persists even when rationales are hidden or nonsensical. 

% \researchfinding{CoT activates a robust ``reasoning mode'' where models leverage generated tokens as external memory to execute stepwise internal computation, including fetching and carrying forward intermediate results and the maintenance of internal states. This core process persists even when CoT rationales are hidden or nonsensical. However, this internal computation is not strictly sequential but a parallel process involving multiple pathways and shortcuts. }

\begin{takeaway}
    \textbf{Takeaway:}
    CoT activates a robust ``reasoning mode'' where models leverage generated tokens as external memory to execute stepwise internal computation, including fetching and carrying forward intermediate results and updating internal states. This core process persists even when CoT rationales are hidden or nonsensical. Yet, this internal computation is not strictly sequential but a parallel process involving multiple pathways and shortcuts.
\end{takeaway}

% \citet{cot_in_the_wild_not_faithful} observed cases of models ``silently correcting'' their intermediate errors, \textit{i.e.}, the model might make a mistake in an early step internally, but still arrive at the correct answer. Similarly, \citet{measuring_Faithfulness_CoT} showed that deleting or perturbing individual CoT steps often has no effect on the answer. 

% \paragraph{Summary.}
% Peering into the activations and attention patterns of CoT-prompted models reveals concrete evidence for the conceptual stages of reasoning. Specialized attention heads retrieve and propagate information between steps; neuron activations encode an intermediate state of the problem; and dedicated "readout" components finally translate this internal state into an answer. These mechanistic insights not only explain how CoT unlocks complex problem-solving but also align with theories of transformers emulating iterative algorithms. While much remains to be understood, and the potential for hidden computations challenges transparency, the emerging picture is that CoT works by aligning the model’s internal architecture with the structure of a stepwise algorithm.

% \subsection{Why Does Chain-of-Thought Outperform Implicit Reasoning?}
% \subsection{Why Can Chain-of-Thought Outperform Implicit Reasoning?}
% \subsection{Why Can CoT Better Elicit Reasoning Abilities in LLMs?}
\subsection{Why CoT Enhances Reasoning Abilities?}

Empirically, explicit reasoning with CoT often solves complex tasks more accurately than implicit latent reasoning. Several reasons have been identified for why CoT prompting dramatically improves reasoning performance. 

% Mechanistic studies have converged on four primary, complementary hypotheses explaining CoT’s superiority: it fundamentally augments the transformer's computational expressiveness, provides an essential external working memory substrate, mitigates reliance on non-compositional statistical shortcuts, and supplies a rich, structured training signal that promotes generalization. 

\paragraph{CoT augments computational expressiveness.}
% The most compelling explanation centers on \textit{computational complexity theory}, which reveals that CoT fundamentally expands the computational power of transformer architectures. 
Recent theoretical studies demonstrate that CoT enhances transformers' expressiveness and computational capacity, enabling them to solve problems in higher complexity classes. A standard transformer decoder without CoT performs constant-depth computation per token, limiting it to the complexity class $\texttt{TC}^0$~\cite{DBLP:conf/nips/MerrillS23,DBLP:journals/tacl/MerrillS23,DBLP:conf/icml/0001CP23}. Such models theoretically cannot solve inherently serial problems because the required computation depth grows with input size, while the model's depth is fixed. CoT breaks this limit. By feeding the output back into the input, CoT allows the transformer to simulate an RNN or a Turing Machine. The effective depth of the computation becomes proportional to the length of the generated chain. This elevates the transformer's expressiveness to Polynomial Time ($\texttt{P}$)~\cite{DBLP:conf/iclr/MerrillS24}, making inherently serial or recursive computations solvable where they otherwise are not~\cite{DBLP:conf/nips/FengZGY0W23,CoT_serial_problems,DBLP:conf/iclr/KimS25,DBLP:conf/icml/BavandpourHR025}. 

% All these results support the notion that Chain-of-Thought effectively increases the reasoning depth of LLMs, making inherently serial or recursive computations learnable where they otherwise were not. 
% Theoretical work by \citet{CoT_serial_problems} shows that CoT strictly increases a transformer's expressiveness, enabling inherently serial computations that standard transformers cannot perform. With CoT, a model can simulate multi-step algorithms that a single forward pass cannot. This translates to markedly higher accuracy on tasks requiring multiple reasoning steps. 

\paragraph{CoT introduces modularity that reduces sample complexity.} CoT decomposes complex tasks into granular, independent sub-problems. This modularity provides an inductive bias that matches the structure of complex, multi-step problems, enabling the model to master tasks with significantly less data. Through both experimental and theoretical evidence, \citet{DBLP:conf/nips/LiSGPO23} demonstrated that CoT decouples in-context learning into a ``filtering'' phase and a ``learning'' phase that significantly reduces the sample complexity required to learn compositional structures like MLPs. Extending this learnability perspective, \citet{DBLP:conf/iclr/Yang0W25} demonstrated that CoT can render inherently ``unlearnable'' tasks efficiently learnable by reducing the sample complexity of the overall task to that of its hardest individual reasoning step. \citet{DBLP:conf/iclr/WenZLZ25} further identified that this efficiency stems from the \textit{sparse sequential dependencies} among tokens. CoT induces interpretable, sparse attention patterns that enable polynomial sample complexity, whereas implicit reasoning requires exponentially many samples to disentangle dense dependencies. 

% They showed that CoT induces interpretable, sparse attention patterns that align with the Transformer's architectural bias, whereas implicit reasoning requires learning exponentially harder dense dependencies. 
% They showed that while implicit learning of functions like parity requires exponentially many samples to disentangle dense dependencies, CoT enables the model to attend to local, sparse relationships, thereby achieving polynomial sample complexity. 

\paragraph{CoT enables more robust reasoning.}
First, evidence has been found that CoT \textit{promotes robust generalization} by encouraging models to learn generalizable solution patterns rather than overfitting to surface-level statistical shortcuts. For example, \citet{yao2025unveiling} demonstrated that CoT-trained models induce a two-stage generalizing circuit that internalizes the reasoning process, leading to strong OOD generalization even in the presence of training noise. Complementing this, \citet{li2025training} provided a theoretical guarantee for CoT generalization, showing that CoT maintains high performance even when context examples are noisy or erroneous, as it relies on step-by-step pattern matching rather than fragile input-output mappings. 
Second, CoT helps \textit{reduce the propagation of errors} during reasoning. \citet{gan2025rethinking} identified a ``snowball error effect'' in implicit reasoning, where minor inaccuracies accumulate into significant failures. They demonstrated that CoT-based strategies mitigate this by expanding the reasoning search space, which effectively lowers the probability of cumulative information loss and prevents errors from cascading through the reasoning chain. 

\begin{takeaway}
    \textbf{Takeaway:}
    CoT enhances LLM reasoning capabilities through three primary mechanisms: 1) It breaks the constant-depth limitation of the standard transformer, extending its effective computational depth. 2) It introduces modularity and inductive bias more aligned with multi-step reasoning, thus reducing the sample complexity required to learn complex tasks. 3) It facilitates robust generalization to OOD data and mitigates error propagation.
%resistance to spurious correlations and  mitigation of error accumulation
\end{takeaway}

% There are also theoretical analyses have shown the usefulness of CoT from the perspective of model expressivity and generalization [16, 59, 49, 69, 116]. By employing CoT, the effective depth of the transformer increases as the generated outputs are fed back into the input [16]. This aligns with our findings, suggesting that a two-layer transformer is sufficient for learning generalizing circuits through CoT training. 

% \section{How does Explicit CoT Training Improve Reasoning?}

% \subsection{How is Chain-of-Thought Capability Learned?}

\subsection{Does Chain-of-Thought Equate to Explainability?}
\label{subsec:explicit:cot_explanation}

Explicit reasoning appears to provide transparency, leading users to assume that CoT explanations accurately reveal how the model arrived at an answer. However, substantial evidence indicates that CoT outputs often do not faithfully reflect the model's actual decision-making process~\cite{measuring_faithfulness,llm_dont_always_say,dont_say_what_they_think-anthropic,cot_is_not_explainability}, a phenomenon referred to as the \textit{unfaithfulness} of CoT reasoning.

\paragraph{Evidence of CoT unfaithfulness.}
Recent studies reveal that CoT frequently functions as \textit{post-hoc rationalization} rather than the causal driver of predictions~\cite{think_to_talk,cot_in_the_wild_not_faithful,analysing_cot_dynamics}. For instance, \citet{llm_dont_always_say} demonstrated that models often alter their predictions based on spurious cues, such as the reordering of multiple-choice options. 
In such cases, the models still tend to confabulate logical-sounding CoT rationales that hide the actual spurious cause of their decision. Similarly, when correct answers are injected as hints, models often invent spurious derivations to support the injected answer without acknowledging the hint's influence~\cite{dont_say_what_they_think-anthropic}. Furthermore, mechanistic analyses uncovered ``silent error corrections'', where models internally correct mistakes without updating the CoT rationale~\cite{cot_in_the_wild_not_faithful}. Unfaithfulness is also evident in \textit{sycophancy}, where models prioritize agreement with user beliefs over truthfulness. Even when models possess the correct internal knowledge, they frequently concede to incorrect user premises and generate plausible rationales to justify these compliant responses~\cite{understanding_sycophancy}. Collectively, these findings highlight a fundamental disconnect between verbalized rationales and internal computations, challenging the premise that CoT equates to explainability. 

% highlight a disconnect between verbalized steps and internal computation: models may commit "silent errors" in the CoT yet still output the correct answer via parallel latent processing~\cite{Arcuschin2025}, or describe algorithmic procedures while actually relying on memorized shortcuts or lookup tables~\cite{Lindsey2025}. 

% finds that CoT explanations can be unfaithful and act as post-hoc rationalizations, especially when models are biased towards incorrect answers. 
% \paragraph{CoT does not reflect spurious reasoning shortcuts.} \citet{llm_dont_always_say} demonstrated that when subtly reordering multiple-choice options in a question, the model's final choice often changed due to positional biases learned by the model. However, the CoT explanation produced by the model looks perfectly rational. The model confabulates a logical-sounding rationale that hides the actual spurious cause of its decision. Recent studies discovered several phenomena related to the unfaithfulness of CoT reasoning. 

% \paragraph{Why do CoT explanations diverge from internal computation?}
\paragraph{Mechanistic understanding of CoT unfaithfulness.} 
Recent mechanistic analyses attribute this unfaithfulness to a fundamental mismatch between the distributed, parallel nature of transformer computation and the sequential nature of explicit reasoning. As discussed in Section~\ref{subsec:explicit:shortcuts}, many works have revealed the \textit{distributed nature} of LLMs' internal reasoning; transformer-based LLMs frequently employ multiple redundant computational pathways to process information, \textit{e.g.}, simultaneously leveraging memorization, heuristics, and algorithmic circuits~\cite{hydra_effect,mechanistic_cot,Arithmetic_Without_Algorithms}. Consequently, CoT only acts as a ``lossy projection'' of high-dimensional internal states, often capturing only a fraction of the model's actual decision process~\cite{mechanistic_cot}. Because computation is highly distributed, a single CoT rationale can capture at most one of many simultaneous causal pathways. As a result, CoTs typically omit influential factors and serve only as partial, post-hoc rationalisations of the model’s underlying distributed, superposed computation~\cite{cot_is_not_explainability}. This architectural dissonance makes unfaithfulness difficult to mitigate. \citet{on_hardness_faithful_cot} demonstrated that even when training objectives explicitly penalize inconsistency, models still revert to plausible-but-not-causal explanations on complex tasks, highlighting the inherent difficulty in eliciting faithful CoT reasoning from LLMs. 

% Because computation is highly distributed, a single CoT rationale can capture at most one of many simultaneous causal pathways, serving only as a partial, post-hoc rationalisation of the model's underlying superposed computation~\cite{cot_is_not_explainability}. 

% [26] How to think step-by-step: A mechanistic understanding of chain-of-thought reasoning
% [27] A mathematical framework for transformer circuits
% [31] Transformer feed-forward layers build predictions by promoting concepts in the vocabulary space
% [42] Language models use trigonometry to do addition
% [47] Measuring faithfulness in chain-of-thought reasoning
% [54] The hydra effect: Emergent self-repair in language model computations 
% [57] Progress measures for grokking via mechanistic interpretability
% [59] Arithmetic without algorithms: Language models solve math with a bag of heuristics
% [62]  In-context learning and induction heads
% [65] Understanding addition in transformers
% [72] On the hardness of faithful chain-of-thought reasoning in large language models

% Consequently, a model may verbalize a step-by-step algorithmic derivation while its actual output is driven by a faster latent shortcut or memorized pattern (Lanham et al., 2023), rendering the CoT a post-hoc rationalization rather than a true causal account. 

\begin{takeaway}
    \textbf{Takeaway:}
    While chain-of-thought offers the appearance of transparency, it does not equate to faithful explainability. CoT often functions as post-hoc rationalization rather than a true reflection of the model's internal processing. Mechanistically, this unfaithfulness stems from a structural mismatch between the distributed, parallel computation of transformers and the sequential nature of explicit reasoning. 
%resistance to spurious correlations and  mitigation of error accumulation
\end{takeaway}

% While Chain-of-Thought offers the appearance of transparency, it does not equate to faithful explainability. Empirical evidence indicates that CoT often functions as post-hoc rationalization, where verbalized steps obscure the true drivers of prediction, such as spurious cues, sycophancy, or latent shortcuts. Mechanistically, this unfaithfulness stems from a structural mismatch: the distributed, parallel computation of transformers cannot be fully captured by sequential verbalization. Consequently, CoT acts as a lossy projection of internal states, creating an inherent barrier to generating reasoning that is causally consistent with the model's internal processing.

\section{Future Research Directions}
% 1. rigorous analysis on real-world scenarios 
% 3. Bridging the Faithfulness Gap
% 3. Latent CoT
% 4. White-box Evaluation 
% 5. From Mechanistic Interpretation to Model Control

\paragraph{Rigorous causal analysis in real-world settings.}
% rigorous analysis on real-world scenarios 
A fundamental challenge in current mechanistic research is \textit{the disparity between idealized experimental settings and the complexities of real-world reasoning}. First, the reliance on toy models and synthetic data limits the generalizability of current findings. For example, while the ``grokking'' phenomenon has been identified as a potential pathway for the emergence of implicit multi-hop reasoning, most empirical evidence is derived from toy models trained from scratch on synthetic tasks (\S~\ref{subsec:implicit:learning}). Consequently, it remains an open question whether the phase transitions observed in these controlled environments truly govern the development of reasoning capabilities in foundation models trained on large-scale, naturalistic corpora. 

Second, the field should move beyond correlational analysis, which only proves information presence, to \textit{rigorous causal verification} within these complex settings. Unlike clean synthetic environments, real-world data is ubiquitous with spurious cues, making it difficult to distinguish genuine reasoning circuits from robust shortcut heuristics (\S~\ref{subsec:implicit:shortcuts}). 
% (\textit{e.g.}, factual co-occurrence). 
Therefore, causal interventions are crucial for proving that identified internal representations are truly drivers of correct inference in the wild. This understanding should ideally translate into \textit{robust training-time interventions} that penalize such shortcuts, forcing models to learn generalizable algorithms despite the noisy data distribution. Ultimately, future work must aim to synthesize these insights into a unified theoretical framework that explains how diverse components, from memorization circuits to reasoning heads, interact within the massive scale of foundation models. 

% Second, the field should move beyond correlational analysis, which only proves information presence, to \textit{rigorous causal interventions} that establish whether specific internal representations are truly necessary and sufficient for correct inference. Furthermore, because real-world data is ubiquitous with spurious cues, distinguishing genuine reasoning circuits from robust shortcut heuristics (\textit{e.g.}, factual co-occurrence) is critical. We currently lack robust training-time interventions that can effectively penalize these shortcuts and force models to learn generalizable reasoning algorithms in the wild. Ultimately, future work must aim to synthesize these isolated mechanistic insights into a unified theoretical framework that explains how these diverse components, from memorization circuits to reasoning heads, interact within the massive scale of foundation models in real-world reasoning. 

% synthetic settings on toy models trained from scratch or indirect analyses of training data, which may not fully reflect how reasoning abilities develop in real-world LLMs. 
% This leaves open questions regarding whether similar phase transitions governing the development of multi-hop reasoning can generalize to models trained on large-scale, naturalistic corpora. 
% To mature the field of mechanistic interpretability, research must bridge the gap between idealized experimental settings and the complexities of real-world LLMs. 

\paragraph{Bridging the faithfulness gap of explicit CoT reasoning.} 
As discussed in \S~\ref{subsec:explicit:cot_explanation}, a critical bottleneck in current LLMs is the ``functional rift''~\cite{mechanistic_cot} between the model's internal, parallel processing and its sequential, explicit CoT reasoning. This structural mismatch forces models to compress high-dimensional, distributed latent states into a low-bandwidth stream of discrete tokens, often resulting in CoT that functions as a post-hoc rationalization rather than a causal driver. To address this, future research must explore \textit{white-box alignment methods} that enforce a causal link between implicit and explicit reasoning. 
Promising avenues include developing training objectives that penalize discrepancies between the model's hidden states (its true decision process) and its generated rationale~\cite{DBLP:journals/corr/abs-2509-23095,wang2025elicitingchainofthoughtbasellms}, imposing architectural constraints that compel the model to rely solely on the generated CoT for subsequent steps~\cite{markovian2024}, as well as ``self-explaining'' dense internal representations into faithful natural language steps~\cite{xnode2025}. 
% Promising avenues include developing training objectives that penalize discrepancies between the model's hidden states (its true decision process) and its generated rationale~\cite{DBLP:journals/corr/abs-2509-23095,wang2025elicitingchainofthoughtbasellms}, as well as imposing architectural constraints that compel the model to rely solely on the generated CoT for subsequent steps~\cite{markovian2024}. Additionally, recent work has investigated mechanisms to ``self-explain'' dense internal representations into faithful natural language steps~\cite{xnode2025}. 
% Further exploration of these directions offers a promising path toward aligning explicit outputs with internal dynamics, ensuring CoT serves as a valid window into the model's computation. 
Further exploration of these directions is critical for aligning explicit outputs with internal dynamics, ensuring CoT serves as a valid window into the model's computation. 

% Together, these approaches offer a promising path toward aligning explicit outputs with internal dynamics, ensuring CoT serves as a valid window into the model's computation. 
% Advancing these research avenues is critical to ensure that CoT functions as a transparent readout of the model's computational trajectory instead of a plausible confabulation. 

% As discussed in previous sections, the "linearization dilemma" forces models to compress high-dimensional, distributed latent states into a low-bandwidth stream of discrete tokens, often resulting in CoT that functions as a post-hoc rationalization rather than a causal driver. 

\paragraph{Mechanistic understanding of Latent CoT reasoning.}
Beyond the dichotomy of implicit and explicit CoT, an emerging paradigm is \textit{latent CoT reasoning}~\cite{latent_cot_survey,implicit_reasoning_survey}, where models are designed to simulate explicit reasoning trajectories entirely within hidden states. Unlike standard implicit reasoning, which relies on the fixed depth of a standard transformer, latent CoT architectures often introduce additional computational capacity via continuous ``thought tokens'', iterative refinement, or recurrent state updates, frequently learning these behaviors by distilling explicit CoT data into latent representations. This approach theoretically offers the best of both worlds: it broadens the model's expressive capacity and computational depth while eliminating the redundant decoding costs of natural language tokens. 
% This approach theoretically offers the best of both worlds: 1) similar to explicit CoT, it broadens the model's expressive capacity and computational depth, and 2) similar to implicit reasoning, it eliminates the redundant decoding costs of natural language tokens. 

While various latent CoT architectures have been proposed~\cite{DBLP:COCONUT,DBLP:CCOT,CODI_paper}, mechanistic interpretability has lagged significantly behind these innovations. While a vast body of work has explored the latent reasoning mechanisms of \textit{standard transformers} (\S~\ref{subsec:implicit:internal_mechanism}), research into the internal dynamics of these novel latent CoT models remains limited~\cite{zhang2024uncoveringlatentcotvector,wang2025latentspace,zhang2025latenttokensthinkcausal}. Critical open questions remain: Does distilling explicit CoT truly force the model to internalize a sequential, step-by-step reasoning process, or does the model collapse the teacher's rationale into high-dimensional statistical shortcuts? Therefore, gaining more mechanistic insights is crucial for designing next-generation latent CoT architectures and training objectives that effectively combine the interpretability of explicit reasoning with the efficiency of implicit computation.

\paragraph{White-box evaluation metrics for LLM reasoning.}
As we gain a deeper mechanistic understanding of multi-step reasoning, it should guide the development of evaluation protocols that go beyond simple end-task accuracy. Current black-box metrics (\textit{e.g.}, final accuracy) are increasingly insufficient, as models frequently arrive at correct answers via non-robust shortcuts, statistical heuristics, or ``bag-of-words'' processing (\S~\ref{subsec:implicit:shortcuts}). To rigorously distinguish genuine reasoning from sophisticated pattern matching, the field requires \textit{``white-box'' evaluation metrics} that integrate model internals into the evaluation protocol. Pioneering efforts have begun to explore this direction. For example, \citet{cao2025modelutilitylawevaluating} introduced a mechanism-interpretable metric (MUI) that quantifies the ``effort'' required to solve a task, defined as the proportion of activated neurons or features. A truly capable model should achieve higher performance with lower effort. While this area remains under-explored, developing metrics that not only score the final output but also verify the presence of necessary internal computational signatures, such as the formation of bridge entities in intermediate layers~\citep{yang2024large_shortcut}, is a crucial future trend. By defining reasoning not just as the correct outcome but as the execution of a verified internal process, we can prevent the overestimation of model capabilities and ensure that improvements on leaderboards reflect true algorithmic generalization. 

\paragraph{From mechanistic interpretation to model control.} 
While current research has successfully identified various reasoning circuits, such as iteration heads or deduction heads, most work remains observational. A major frontier for future study is the shift towards \textit{pragmatic interpretability}~\cite{nanda2025pragmatic,nanda2025interpretability}, moving from passively explaining mechanisms to actively leveraging them for model control and editing, a paradigm closely aligned with \textit{Representation Engineering (RepE)}~\cite{DBLP:journals/tmlr/WehnerATKF25}. For example, if we can reliably identify the specific components responsible for multi-step logic, \textit{e.g.}, the state-maintenance neurons identified by \citep{DBLP:conf/acl/RaiY24}, we can potentially intervene in real-time to correct reasoning errors or suppress shortcut neurons~\citep{ju2024investigating}. 
% This opens the door to developing ``self-correcting'' architectures that monitor their own internal activation patterns and dynamically modulate their computation paths when a shortcut or ``silent error'' is detected, thereby enhancing the reliability and safety of reasoning systems. 
Such interventions enable the development of ``self-correcting'' architectures that actively monitor internal states to detect and resolve failures like ``silent errors'' on the fly. 
% utilize internal monitors to detect deviations from faithful reasoning paths (\textit{e.g.}, “silent errors” or heuristic overuse) and dynamically modulate their own computation. 
% By closing the loop between understanding and control, interpretability transforms from a post-hoc analysis tool into a foundational component for building reliable and safe reasoning systems. 
Ultimately, this enables a transition from interpretability as a passive analysis tool to an active, foundational component for robust and safe reasoning systems.

% While current research has successfully mapped various reasoning circuits—such as iteration heads or deduction heads—most work remains observational. A major frontier for future study is moving from interpreting these mechanisms to active model control and editing. 
% If we can reliably identify the specific components responsible for multi-step logic (\textit{e.g.}, the state-maintenance neurons identified by \citep{DBLP:conf/acl/RaiY24}), we can potentially intervene in real-time to correct reasoning errors or suppress shortcut neurons~\citep{ju2024investigating}. 
% This direction holds the promise of developing "self-correcting" architectures that monitor their own internal activation patterns and dynamically modulate their computation paths when a shortcut or "silent error" is detected, thereby enhancing the reliability and safety of reasoning systems.

\section{Conclusion}
In this survey, we provided a comprehensive overview of the mechanisms underlying multi-step reasoning in large language models. We structured our analysis around two fundamentally distinct computational paradigms: implicit reasoning and explicit reasoning. Through a framework of seven interconnected research questions, we systematically explored the internal dynamics of latent inference, the emergence of reasoning capabilities, and the mechanistic impact of chain-of-thought prompting on model computation and expressiveness. Despite significant progress in opening the black box, critical challenges remain. 
% We highlighted the persistent "functional rift" between parallel internal states and sequential explicit outputs, which complicates the faithfulness of generated explanations. 
Looking ahead, we outlined a roadmap for future research, emphasizing the necessary shift from passive observation to causal intervention and the need for rigorous verification in real-world settings to build more reliable reasoning systems. 

\section*{Limitations}
This survey concentrates strictly on the mechanistic understanding of multi-step reasoning within transformer-based LLMs. Consequently, we do not cover other aspects of reasoning, such as probabilistic inference, creative planning, or commonsense reasoning, which may operate under different mechanistic principles. Additionally, our scope is limited to the current paradigm of text-based transformers; we do not extensively address reasoning mechanisms in Multimodal LLMs (MLLMs), alternative architectures like Diffusion Language Models (DLMs), or neural networks that predate the modern era of large language models. 

% the broader history of symbolic AI that predates the modern era of large language models. 

% \section*{Acknowledgments}

\bibliography{custom}

% \appendix
% \section{Example Appendix}
% \label{sec:appendix}
% This is an appendix.

\end{document}